\documentclass[10pt,journal,compsoc]{IEEEtran}
\ifCLASSOPTIONcompsoc
  \usepackage[nocompress]{cite}
\else
  \usepackage{cite}
\fi
\ifCLASSINFOpdf
  \usepackage[pdftex]{graphicx}
  \graphicspath{{Images/}}
  \DeclareGraphicsExtensions{.pdf,.jpeg,.png}
\else
\fi
\usepackage{amsmath}
\usepackage{amsfonts}
\usepackage{amssymb}
\usepackage{booktabs}
\usepackage{threeparttable}
\usepackage{array}
\usepackage{multirow}
\usepackage{color, xcolor}
\begin{document}
%
\title{Watch Out for the Confusing Faces:\\Detecting Face Swapping with the Probability Distribution of Face Identification Models}

\author{Yuxuan~Duan, Xuhong~Zhang, Chuer~Yu, Zonghui~Wang, Shouling~Ji, Wenzhi~Chen %

\IEEEcompsocitemizethanks{
\IEEEcompsocthanksitem Y. Duan, X. Zhang, C. Yu, Z. Wang, S. Ji and W. Chen are with the College of Computer Science and Technology, Zhejiang University, Hangzhou, Zhejiang 310027, China; 
S. Ji is also with the school of Electrical and Computer Engineering, Georgia Institute of Technology, Atlanta, Georgia, 30332, USA. \protect\\ 
Email: \{duanyuxuan, zhangxuhong, yuchuer, zhwang, sji, chenwz\}@zju.edu.cn %
}}

%
%

\markboth{Journal of \LaTeX\ Class Files,~Vol.~14, No.~8, March~2023}%
{Shell \MakeLowercase{\textit{et al.}}: Bare Demo of IEEEtran.cls for Computer Society Journals}
%



\IEEEtitleabstractindextext{%
\begin{abstract}
Recently, face swapping has been developing rapidly and achieved a surprising reality, raising concerns about fake content. As a countermeasure, various detection approaches have been proposed and achieved promising performance. However, most existing detectors struggle to maintain performance on unseen face swapping methods and low-quality images.
Apart from the generalization problem, current detection approaches have been shown vulnerable to evasion attacks crafted by detection-aware manipulators.  
Lack of robustness under adversary scenarios leaves threats for applying face swapping detection in real world.
In this paper, we propose a novel face swapping detection approach based on face identification probability distributions, coined as IdP\_FSD, to improve the generalization and robustness.
IdP\_FSD is specially designed for detecting swapped faces whose identities belong to a finite set, which is meaningful in real-world applications. Compared with previous general detection methods, we make use of the available real faces with concerned identities and require no fake samples for training.
IdP\_FSD exploits face swapping's common nature that the identity of swapped face combines that of two faces involved in swapping. We reflect this nature with the confusion of a face identification model and measure the confusion with the maximum value of the output probability distribution.
What's more, to defend our detector under adversary scenarios, an attention-based finetuning scheme is proposed for the face identification models used in IdP\_FSD.
Extensive experiments show that the proposed IdP\_FSD not only achieves high detection performance on different benchmark datasets and image qualities but also raises the bar for manipulators to evade the detection.
\end{abstract}

}

\maketitle

\IEEEdisplaynontitleabstractindextext

%
\IEEEpeerreviewmaketitle

\IEEEraisesectionheading{\section{Introduction}\label{sec:introduction}}

\IEEEPARstart{F}{ace} forgery technologies are constantly evolving and becoming increasingly realistic. One major category of face forgery is face swapping \cite{chen2020simswap, li2020advancing, zhu2021one, xu2022high, xu2022styleswap}, which aims at replacing the face of one person in an image or a video with the face of another person \cite{tolosana2020deepfakes}. Despite its applications in fields like entertainment, face swapping can cause serious social risks. For example, the manipulator may swap the face of politicians or celebrities into any video to forge a speech they never spoke, misleading the public.

To tackle the malicious application of face swapping, various detection methods have been proposed. A common way is to model face swapping detection as a binary classification task and collect existing face forgery images to train the classifier \cite{afchar2018mesonet, Li_2019_CVPR_Workshops, rossler2019faceforensics++, qian2020thinking, li2020face, zhao2021multi, chen2021local, wang2021representative, cao2022end, kong2022detect}. Although these approaches achieve high detection accuracy on benchmark datasets, they still struggle to meet some real-world demands. In particular, there remain three key challenges. 
First, most existing detectors have trouble in detecting fake samples generated by unseen face swapping methods\cite{cozzolino2021id}. It is a severe problem since the methods for generating forgery faces are updating fast and are usually unknown to the detector. 
Second, current methods often suffer a performance drop on images with degradation\cite{woo2022add}, which are common in the spread of forgeries. It is because they rely on low-level artifacts such as blur at blending boundary\cite{li2020face}, which are prone to image degradation.
Finally, most of the previous works assume the manipulator knows nothing about the detection mechanism. However, it is highly possible for a malicious manipulator to infer how the detection works and polish their fake images to evade detection. Actually, researches have shown face forgery detectors are vulnerable to adversarial examples just as other DNN-based classifiers \cite{carlini2020evading, neekhara2021adversarial, hussain2021adversarial, jia2022exploring}. Thus, it is necessary to develop face forgery detection robust in adversarial settings.

In addition to the above challenges, there is another practical situation neglected by most existing detection methods, where defenders only care about detecting forgery faces of people in a finite set. For example, an enterprise may be concerned about their spokesman's fake videos, and a real-name social platform like LinkedIn may be expected to prevent their registered users' forgery images from spreading on the platform. 
In these situations, detection methods with enhanced generalization and robustness yet limited to specified identities can be useful. 
While the previous general face forgery detection also applies to such scenarios, they rarely utilize the available real portraits of the identities of concern, which we show helpful in solving the challenges. 
There have been a few works \cite{cozzolino2021id, dong2022protecting, jiang2021practical, ramachandran2021experimental} that make use of reference real images or videos to improve generalization. However, these approaches only leverage one specific reference and for inference only, making the detection accuracy sensitive to the choice of the reference.
For example, some works classify a query image as fake if its distance to the reference with the same identity is larger than a threshold \cite{dong2022protecting, ramachandran2021experimental}. However, a large difference in head pose between a real query and its reference can also increase the distance, resulting in false positives. In contrast, we propose to learn stable features for concerned identities by making full use of available real images in the training stage.

In this paper, we present IdP\_FSD, a novel face swapping detection method based on the probability distribution of face identification models, focusing on addressing the aforementioned challenges for a finite set of identities. The key idea is that \textit{the identity of the swapped face is a combination of the two identities involved in face swapping, resulting in more confusion for the face identification models}. Such identity combination results from the basic demand of face swapping, i.e., preserving visual features of both faces, as well as the great challenge to completely disentangle identity with other facial attributes\cite{xu2022high, chang2021learning}.
And we measure how confusing one face image is to the model with the prediction probability distribution\cite{hendrycks17baseline}. 
To be specific, we first train a face identification model on real faces from the finite set of people whose fake images are in concern. Then in the inference stage, we take the maximum class probability from the softmax distribution and use its magnitude to detect whether a face is swapped or not.
By training on real faces only, our detector requires no prior knowledge about the face swapping methods, thus generalizing well across different methods. Besides, we improve the robustness to image degradation by relying on high-level identity information instead of low-level artifacts.

Moreover, we strengthen our approach for the adversary scenario where the manipulator is aware of our detection principle. It is the face identification model used in the approach that plays an important role. As a baseline, publicly available pre-trained face identity extractors like ArcFace\cite{deng2019arcface} are used to extract identity features, and then a classifier is trained on these features, forming the identification model. The baseline method achieves high forgery detection accuracy in the defense-unaware setting.
In the defense-aware grey-box setting, however, we design an evasion attack and achieve a high evasion rate. It is not surprising since the manipulator can also use the pre-trained identity extractors to disguise their fake images as real in the identity feature space. And experiments show that simply ensembling or finetuning the identity extractors helps little on attack defense.
Therefore, to improve the robustness against grey-box evasion attacks, we design a special attention-based finetuning scheme. Compared with simple finetuning, our method explicitly enlarges the attention area of the face identification model. In this way, the attacks carried out on surrogate identity extractors get harder to transfer to our finetuned identification model.

The main contributions can be summarized as follows.
\begin{itemize}
    \item We propose a novel method to detect face swapping images according to the confusion in the probability distribution of a face identification model named IdP\_FSD.
    The method trains on real images of the concerned identities and makes use of high-level identity features, thus generalizing well across different face swapping methods and image qualities.
    \item We study the robustness in the defense-aware grey-box setting, which is barely considered in previous face swapping detection works. We present an attention-based face identification model finetuning scheme to strengthen IdP\_FSD against grey-box evasion attacks.
    \item Extensive experiments show the proposed IdP\_FSD achieves high detection performance on different forgery datasets. Moreover, the proposed finetuning scheme can remarkably downgrade the attack success rate of evasion attacks applied to fake images.
\end{itemize}

\section{Related Work}

\textbf{Face Swapping Generation.} Face swapping technologies have been receiving great attention and developing continuously for their rich applications. The aim of face swapping is to transfer the identity of a source face into a target face while preserving the attributes like expression, head pose and background of the target face\cite{chen2020simswap}. Earlier face swapping methods are based on 3D models, such as FaceSwap\cite{FaceSwap}, which usually incur obvious artifacts and high computation cost. Recently, face swapping based on deep learning models greatly improve the visual quality and efficiency, enabling its wide use.

Deep learning based face swapping methods are basically built on encoder-decoder architectures. We roughly divide existing face swapping methods into three categories according to the different training designs of the encoder and decoder.
1) Subject specific methods train the encoder-decoder for each pair of source and target subjects, such as Deepfakes\cite{Deepfakes} and Faceswap-GAN\cite{Faceswap-GAN}. 
2) To address the subject specific limitation, researchers proposed to train general face encoder and decoder together for face swapping.
FaceShifter\cite{li2020advancing} utilizes an identity encoder and an attribute encoder to encode the source and target face, then fuses them in each layer of the decoder to generate the swapped face. 
SimSwap\cite{chen2020simswap} follows a similar pipeline but fuses the identity and attribute feature in the latent space before being sent to the decoder, achieving state-of-the-art performance. And there are similar works which improves either the feature fusion \cite{gao2021information} or the identity extraction\cite{kim2022smooth}. Although this line of work has achieved satisfying performance in identity transferring and attribute preserving, the resolution of the face swapping images is still limited.
3) To generate high-resolution swapped faces, GAN inversion based face swapping methods are proposed\cite{zhu2021one, xu2022high, luo2022styleface}. For example, MegaFS\cite{zhu2021one} uses the pre-trained StyleGAN2\cite{karras2020analyzing} as the decoder and trains the encoder only.

With the rapid evolution mentioned above, state-of-the-art face swapping methods have achieved surprisingly high quality, making it difficult to detect them with low level artifacts. It is worth noting that the face blending step which used to be common in early deepfake methods is becoming less necessary. Especially for the methods based on the large-scale GAN with strong capability, generating the whole swapped faces directly is possible. On the other hand, almost all the methods focus on keeping the source identity and target attributes at the same time, which means the combination of two faces is a nature of swapped faces.

\textbf{Face Swapping Detection.}
To tackle the malicious use of face swapping, numerous detection methods have been proposed. Most existing works model face swapping detection as a binary classification problem and focus on designing better the features classified on\cite{Li_2019_CVPR_Workshops, li2020face, kong2022detect, qian2020thinking, chen2021local, li2023forensic}, the classifier network\cite{afchar2018mesonet, rossler2019faceforensics++, zhao2021multi, cao2022end, zhuang2022uia, miao2023f} or the training policies\cite{wang2021representative, sun2022dual, liang2022exploring} to improve the accuracy and generalization.
For example, low-level artifacts are used in early detection methods, such as face warping artifacts\cite{Li_2019_CVPR_Workshops}. To detect fake images with less artifacts, features like frequency\cite{qian2020thinking}, local differences\cite{chen2021local} and symmetry features\cite{li2023forensic} are explored.
As for the network, powerful backbones like Xception\cite{rossler2019faceforensics++} and EfficientNet\cite{zhao2021multi} are usually used. Recent works like UIA-VIT\cite{zhuang2022uia} utilize Transformer in face forgery detection and achieve high performance.
Besides, different training policies such as contrastive learning\cite{sun2022dual} are applied to make the classifier generalize better on unseen face swapping methods. RFM\cite{wang2021representative} encourages the detector to mine more forgery clues to avoid overfitting by occluding the most sensitive face areas during training. And Liang \textit{et al.}\cite{liang2022exploring} proposes an embeddable training framework to disentangle content information and forgery traces, so that the detector can focus on forgery related artifacts.

Despite the numerous efforts, detection methods based on binary classifiers still show limited generalization. Therefore, we turn to train on real faces only to avoid overfitting on fake images generated by specific face swapping methods.
One of the most related works is ICT\cite{dong2022protecting}. ICT detects the identity inconsistency between the inner and outer face caused by the blending step. A transformer is trained on real faces to extract the inner and outer identity embedding simultaneously. Then the distance between two identity embeddings is used as the metric to detect fake images. When assisted with reference real portraits, ICT-Ref achieves state-of-the-art accuracy and generalization.
However, the generalization ability of ICT relies on the existence of the face blending step, which does not exist in some new face swapping methods like SimSwap, since they are capable of generating the whole swapped face. Instead, we exploit the identity combination nature of face swapping.
Another related work is presented in \cite{ramachandran2021experimental}, which calculates cosine similarity between identity features of the probe and a reference real face, then classifies the probe whose cosine similarity is smaller than a threshold as deepfake. We show this method is sensitive to the choice of the reference.
Besides, there are methods that utilize reference real faces to boost the binary classifier such as DISC\cite{jiang2021practical}. It extracts feature maps from reference real faces of the same person to guide the detector's attention.

\textbf{Adversarial Threats in Face Swapping Detection.}
It has been shown that DNNs are vulnerable to adversarial attacks. Therefore, it is reasonable that DNN-based face swapping detectors also share such vulnerabilities.
Hussain \textit{et al.}\cite{hussain2021adversarial} evaluated the deepfake detectors' vulnerability to adversarial examples in white-box and black-box settings, highlighting the threats.
Neekhara \textit{et al.}\cite{neekhara2021adversarial} conducted adversarial attacks on deepfake detectors with different classifier networks. They verified the effectiveness of transfer attacks and universal attacks on deepfake detectors. 
Compared with adversarial attacks in traditional image classification tasks, the perturbation required to fool a deepfake detector is even smaller\cite{carlini2020evading}, probably resulting from the visual similarity between the real and fake faces.
Moreover, attacks specially designed for deepfake detectors have been proposed and show better invisibility\cite{jia2022exploring}.
On the other hand, few detection works pay attention to strengthening the detectors against attacks. As a security-related task, face swapping detection is expected to be more robust against adversarial attacks. Our work considers the grey-box scenarios and makes attempts to increase the difficulty of adversarial attacks.

\textbf{Face Identification.}
Face identification refers to the system that returns the predicted identity of a test image\cite{tong2021facesec}, which is usually modeled as a multi-class classification problem. Similar with face verification, the core of face identification is identity extractors that map faces to embeddings with small intra-identity and large inter-identity distance\cite{deng2019arcface}. There have been several outstanding identity extractors such as FaceNet\cite{schroff2015facenet}, SphereFace\cite{liu2017sphereface}, CosFace\cite{wang2018cosface} and ArcFace\cite{deng2019arcface}, enabling the wide use of face identification. 

In this work, we show face identification models are useful in representing the identity combination feature of swapped faces. Besides, we design an attention-based finetuning scheme for the face identification model to defend the adversarially post-processed swapped faces.

\section{Method}
We present our face swapping detection method IdP\_FSD in this section. The basic idea behind our method is to explore what face swapping images have in common: \textit{the combination of two identities}. We start by presenting the evidence for our intuition. Then we describe the IdP\_FSD framework in detail. Finally, to defend grey-box evasion attacks, we design a finetuning scheme for the face identification models used in IdP\_FSD.

\begin{figure}[!t]
\centering
\includegraphics[width=2.0in]{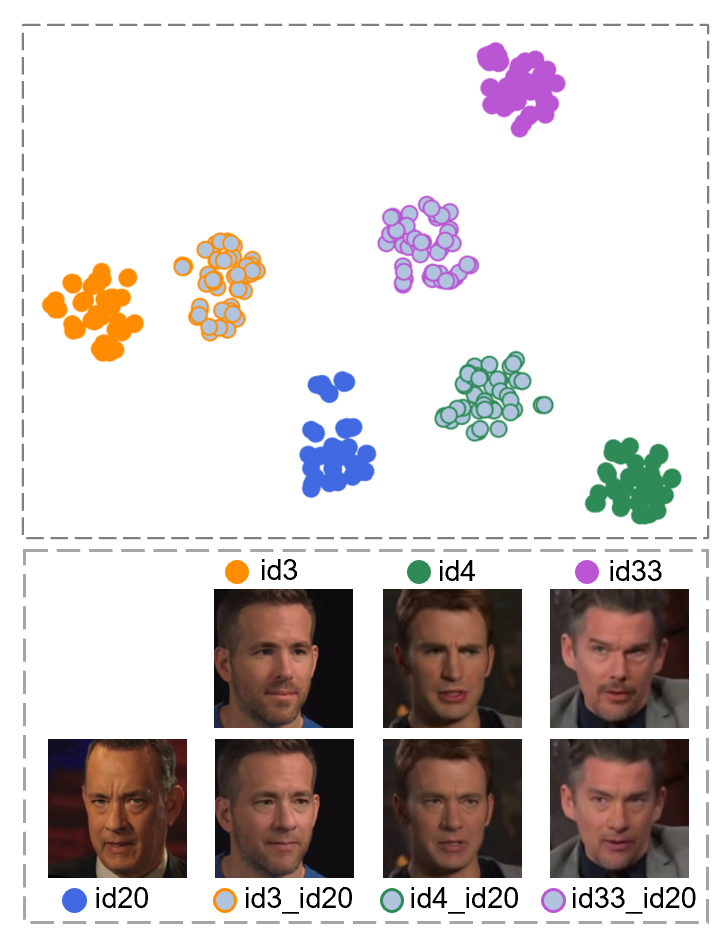}
\caption{The t-SNE\cite{van2008visualizing} visualization of identity embeddings of face swapping images and their source and target real faces. One source identity, i.e., \textit{id20}, is swapped onto three different target faces, resulting in three swapped face groups.}
\label{fig_sim}
\end{figure}

\subsection{Identity Combining Nature of Face Swapping} \label{c3s1}

In this section, we explain and verify the intuition behind our face swapping detection method. Given two faces, face swapping technologies transfer the identity from the source image to the target image while keeping the attributes of the target face\cite{zhu2021one, chen2020simswap}. Ideally, the identity of the swapped face should be identical to that of the \textit{source} face. In fact, the recent face swapping models can make these two identities similar enough for human eyes or metrics like ID retrieval\cite{zhu2021one, tariq2022real}. However, we still observe a discrepancy between the identity of swapped faces and their sources.

Fig. \ref{fig_sim} demonstrates a representative example of such identity discrepancy. We take real images of four identities and fake images with one of them, i.e., \textit{id20}, as the source and the other three as the targets, from CelebDF\cite{li2020celeb}. Then their identity embeddings are extracted with a pre-trained ArcFace\cite{deng2019arcface} and visualized using t-SNE\cite{van2008visualizing} in Fig. \ref{fig_sim}. From the figure we can tell that there exists a gap between the identities of the swapped faces and their sources. Moreover, identities of swapped faces change when transferring to different target faces and tend to lie between the identities of their sources and target. Thus, it is possible to discriminate fake faces from real ones from identity features.

\begin{figure}[!t]
\centering
\includegraphics[width=3.5in]{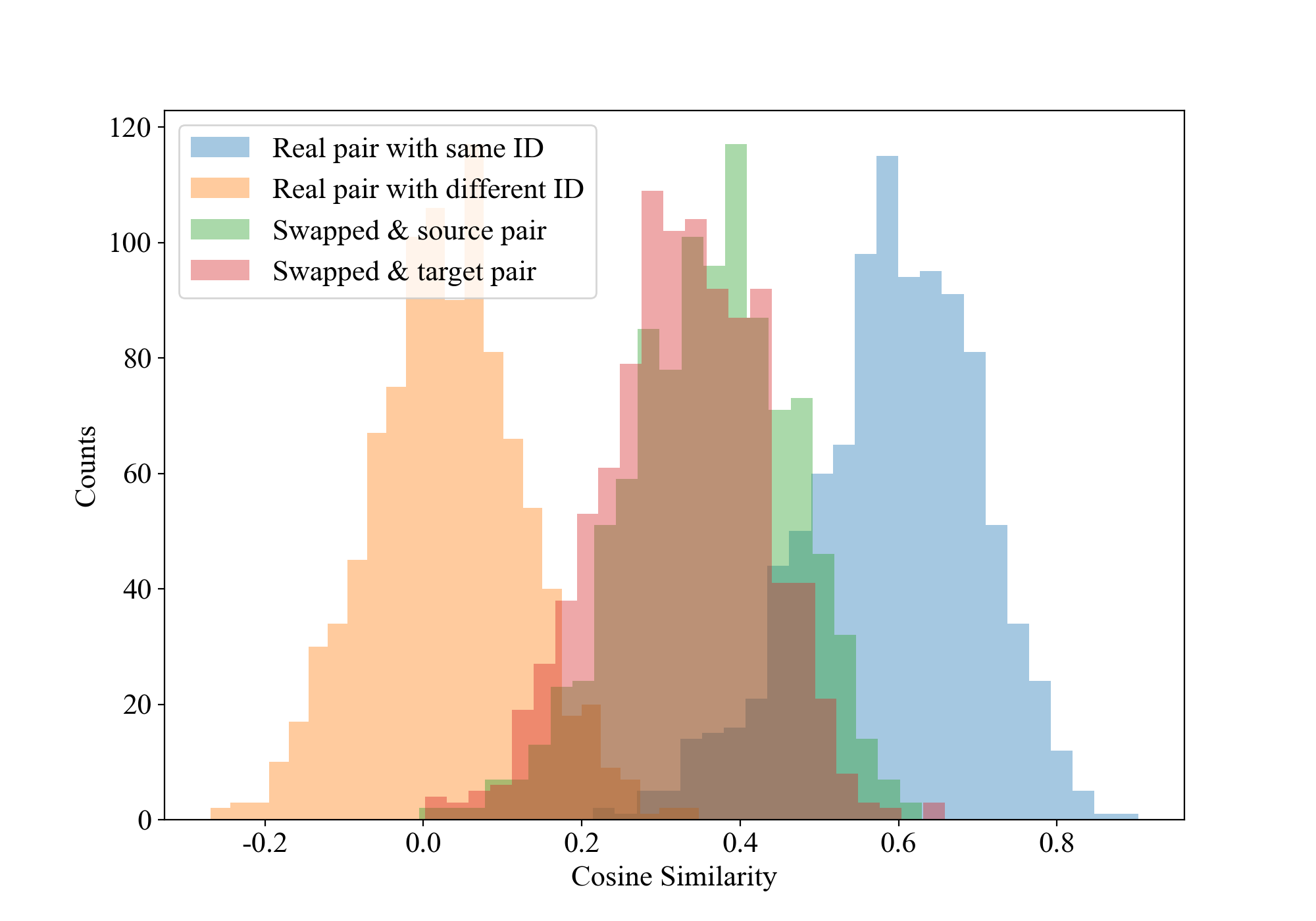}
\caption{The cosine similarity of four kinds of image pairs. (1)two real images with the same identity; (2)two real images with different identities; (3)one fake image and one real image with the same identity as the source of the fake; (4)one fake image and one real image with the same identity as the target of the fake.}
\label{fig_sim_dist}
\end{figure}

To verify the above observation quantitatively, we further measure the similarity between the identity features of swapped faces and their related real faces. The result is displayed in Fig.\ref{fig_sim_dist}. In particular, we randomly sample 1,000 fake images from CelebDF. For each fake image, we sample two real images to form two pairs, one with the source identity and the other with the target, and calculate the cosine similarity for each pair. To make a better illustration, we also plot the similarities of real image pairs with the same identity and different identities as a baseline.
From the figure we can draw following conclusions. 1) The identity of the swapped image is similar with not only its source but also its target, since the green and red distributions are close. That's to say, instead of being identical to the source, \textit{the identity of the swapped face combines that of the source and target}.
2) Compared with a real pair with different identities, the similarity between a swapped face and its source is much higher. Therefore, the swapped can be recognized as the source identity in tasks like face verification, which is consistent with previous works\cite{tariq2022real, li2022seeing}.
However, the similarity is still lower than that of a real pair with the same identity, leaving chances for detection.

In conclusion, the face swapping images combines the identity features of their source and target faces instead of completely maintaining the source identity as expected.
We attribute the phenomenon to the entanglement of the facial attributes and identity features\cite{xu2022high, chang2021learning}. As it's difficult to disentangle attributes and identities, the identity of the target face will inevitably exist in the swapped face.
Thus, the identity combination feature comes from face swapping's main purpose to keep the source identity and target attribute at the same time. We design IdP\_FSD based on this common identity feature to achieve high generalization.

\subsection{Detection System Overview} \label{sec_overview}
\begin{figure}[!t]
\centering
\includegraphics[width=3.0in]{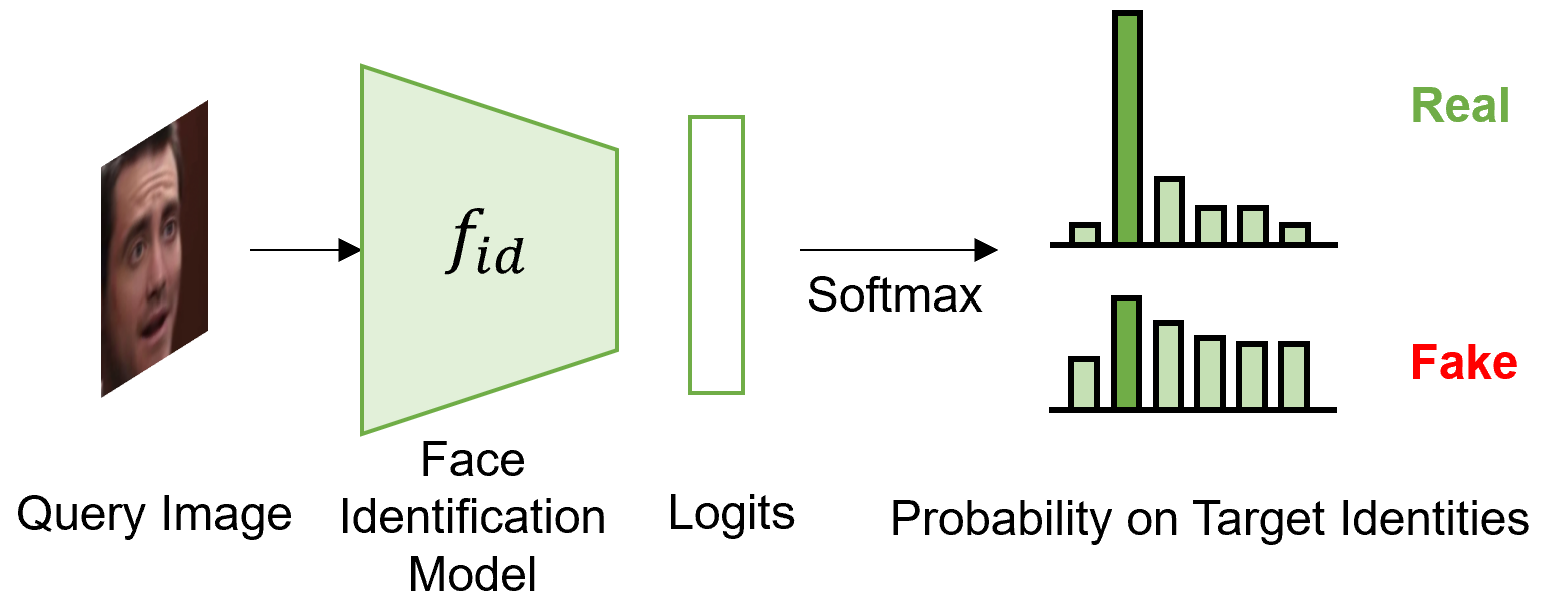}
\caption{Overall detection pipeline of IdP\_FSD. The query image is sent to a face identification model trained on the specific identity set. The maximum value of the probability distribution is used to detect face swapping images.}
\label{fig_method_overview}
\end{figure}

The overall face swapping detection pipeline of IdP\_FSD is displayed in Fig. \ref{fig_method_overview}. Face identification models are used to exploit the aforementioned identity combination nature of face swapping images. Given a query face image $x$, a trained face identification model $f_{id}$ can predict the probabilities of $x$ belonging to each identity in the specific identity set.
For a swapped face that combines two different identities, the model will get confused and output an ambiguous prediction. While for a real image whose identity belongs to the training identity set, the prediction probability distribution will have less confusion.
Therefore, we measure how confusing or surprising an input face is for the face identification model to detect swapped faces.

To be specific, we first train a face identification model on real face images of the concerned identity set. As for the training we adopt two training strategies. One is a standard transfer learning policy with pre-trained identity extractors frozen and the other is a finetuning scheme to defend evasion attacks, which is presented in section \ref{c3s3}.
Then in the detecting stage, we calculate some metrics based on the probability distribution output by the identification model to measure the confusion. 
The actual metric our method selects is the maximum of the probability. The lower the maximum probability is, the more confusing the model gets, thus the more chance the input is a swapped face.

\subsection{Defending Grey-box Evasion Attack} \label{c3s3}
IdP\_FSD performs well on fake images generated by manipulators without any knowledge about the detection principle. However, it is possible for the manipulators to be aware of the detection method and craft their fake images to evade detection, which we refer to as \textit{evasion attack}.
On the other hand, the white-box attack setting where the manipulator not only knows how the detector works, but also has access to the training set and parameters, has been shown difficult to defend\cite{meng2017magnet} and less practical for the manipulator. Therefore, in this work, we study the grey-box evasion attacks against our IdP\_FSD and propose a face identification model finetuning scheme to improve the robustness in grey-box setting, named as IdP\_FSD-ft.

\subsubsection{Grey-Box Threat Model} \label{grey_box_threat_model}
We first define the grey-box threat model in this section, as well as a possible evasion attack against IdP\_FSD.

\textbf{Manipulator's goals.}
In a grey-box setting, the manipulator is aware of the detection method. The goal of the manipulator is to post-process the fake image to make our detector recognize it as a real one. In the meanwhile, the fake image should keep its visual quality and face swapping result. In terms of attacking our detection approach, the specific goal is to make the maximum face identification probability of the fake image as high as possible, so that it is indistinguishable from real images.

\textbf{Manipulator's knowledge and capability.}
A manipulator in the grey-box setting knows our approach detects face swapping with the probability of a face identification model. However, he or she has no access to the parameters or the training set of the face identification model. What he or she can have is the real images with the identity to be swapped which are necessary for generating the fake image. Besides, considering the development of face recognition technology, a manipulator can also collect off-the-shelf pre-trained face recognition models to deploy the attack.

We consider a straightforward attack under the defined grey-box setting. The final goal of the attack is to maximize the maximum identity probability of the fake image $x_f$ as
\begin{equation}
    x_{f}^{adv} = \mathop{\arg\max}_{x^{\prime}:||x^{\prime} - x_f||_p < \epsilon}{\max_i f_{id}(x^{\prime})_i}
\end{equation}

where $\epsilon$ is a small constant. Since the parameter of the face identification model $f_{id}$ used by the detector is not accessible, the adversary needs to find or train a surrogate model. And since the training set is not known either, the classification layer cannot be the same. 

One choice is to attack on the identity feature which is the input of the classification layer. By making use of open-source face recognition models, the manipulator can minimize the distance between the identity features of the fake and its real pair with adversarial attacks as
\begin{equation}
    x_{f}^{adv} = \mathop{\arg\min}_{x^{\prime}:||x^{\prime} - x_f||_p < \epsilon}{d(z(x_f), z(x_r))}
\end{equation}
where $z(x) \colon \chi \to \mathbb{R}^d $ denotes a face recognition model that extracts identity feature in $\mathbb{R}^d$ for an input face. Since the identity extractor used in detection is not known to the manipulator, the manipulator can attack multiple identity extractors simultaneously as
\begin{equation} \label{adv_eq}
    x_{f}^{adv} = \mathop{\arg\min}_{x^{\prime}:||x^{\prime} - x_f||_p < \epsilon}{\sum_{k=0}^K d(z_k(x_f), z_k(x_r))}
\end{equation}

Since the identity feature extractor is frozen in the straightforward version of IdP\_FSD, adversarial fake images would be classified as real as long as their identity features are close enough. 
As a result, the evasion attacks above achieve high evasion rates in grey-box setting.
Simple defenses such as ensembling multiple identity extractors or directly finetuning the extractor only make the evasion rate only drop a little.
Therefore, we design an attention-based finetuning scheme for the face identification model to defend the grey-box evasion attack.

\subsubsection{Attention-based Finetuning}
We attribute the success of the grey-box evasion attack to the similarity of attention between different face recognition models. Without access to the identification models used for detection, the manipulator pulls the swapped and real images close using the surrogate identity extractors.
Thus, the adversary noise is mainly added to the pixels that contribute much to the surrogates.
At the same time, it has been shown that common face identification models tend to pay attention to similar discriminative face areas, such as eyes and noses\cite{dong2022protecting}.
Thus the adversarial noise can transfer among different identity extractors, making IdP\_FSD vulnerable to grey-box evasion attacks.
It is worth noting that the evasion attack is different from the traditional impersonate attacks. Impersonate attacks make a visually unrelated face classified as a specific identity, while the swapped face is visually similar to the source identity and only needs to be pulled closer.

Therefore, to defend our approach against grey-box evasion attacks, we propose to finetune the face identification model to enlarge its attention area explicitly. We adopt two finetuning techniques to realize the aim: attention-based mask and label smoothing.

\textbf{Attention-based mask}. The intuition is to mask the area with the most attention and train the model to predict the identity of the masked face. We follow \cite{wang2021representative} to generate the attention mask according to the gradients. And the model used for generating gradient maps is updated every few epochs for training stability.

\textbf{Label smoothing}. We find using one-hot hard labels for masked training images makes the face identification model overconfident with partial faces. As a result, some swapped faces also get high prediction probability and the detection performance drops. To avoid the overconfidence, label smoothing is introduced by merging the one-hot label and the normal distribution as
\begin{equation}
    y_i = 
    \begin{cases}
    1 - \alpha + {\alpha}/{K} &, i = y_{target} \\
    {\alpha}/{K} &, otherwise.
    \end{cases}
\end{equation}
where $K$ is the number of identities and $\alpha \in \left(0,1\right)$ controls the ratio of one-hot label and normal distribution.

\begin{figure}[!t]
\centering
\includegraphics[width=3.0in]{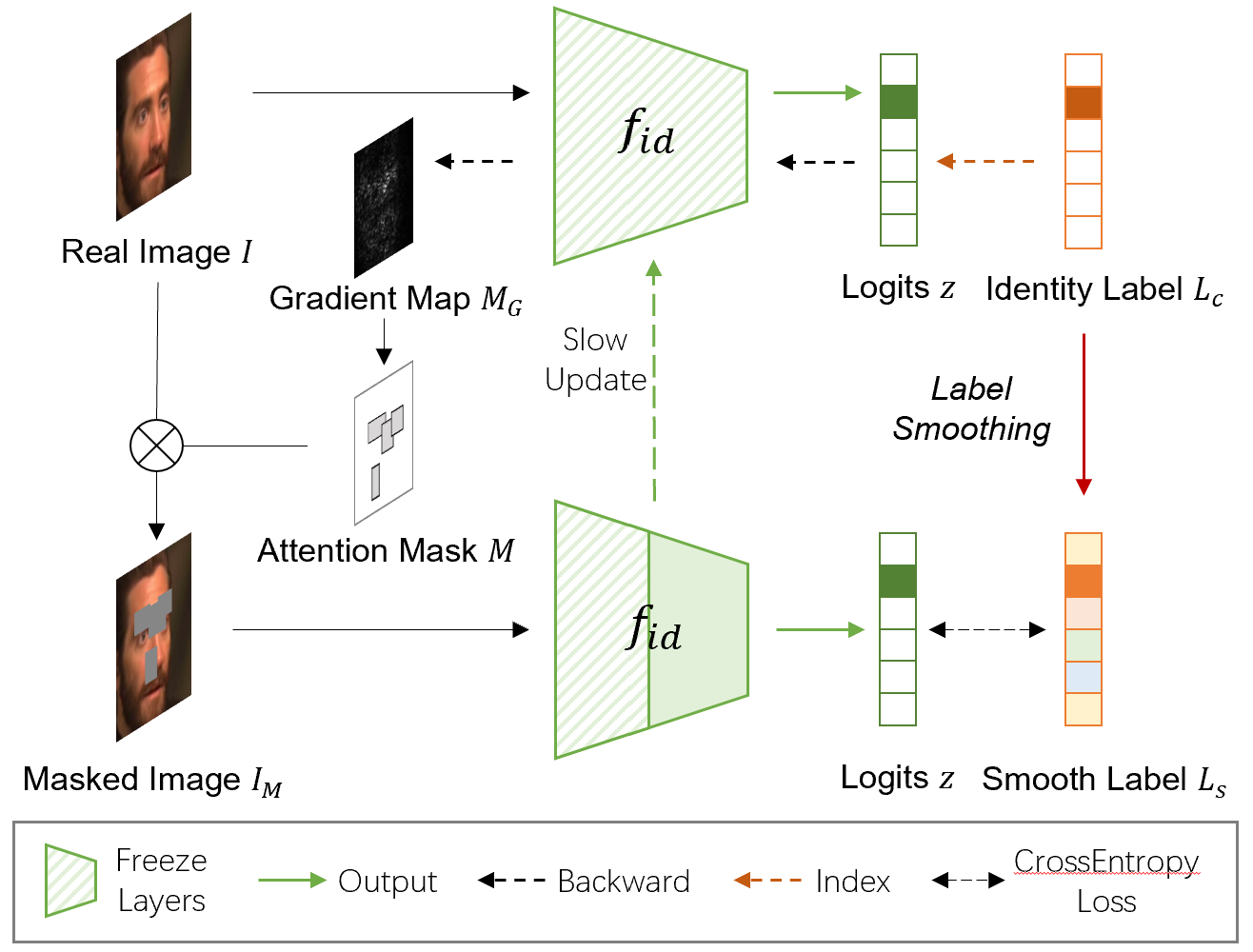}
\caption{The attention-based finetuning scheme. During training, the face area with the most attention is masked to force the model to learn identity from more face areas. The masked image is supervised with smooth labels to avoid overconfidence.}
\label{fig_finetune}
\end{figure}

\section{Experiment}
\subsection{Implementation and Database}
\subsubsection{Dataset}
We evaluate our method on four conventional benchmark deepfake datasets.

\textbf{FaceForensics++ (FF++).} FF++\cite{rossler2019faceforensics++} is a widely used dataset containing 1000 original videos and fake videos obtained by four face manipulation methods: DeepFakes (DF), Face2Face (F2F), FaceSwap (FS) and NeuralTextures (NT). Fake videos generated by FaceShifter (FSh) are added later. Among the fake videos, DF, FS and FSh belong to the face swapping manipulation, while F2F and NT belong to the face reenactment which is out of our scope. Thus, we take our experiment on DF, FS and FSh only. As for the identities of original videos, we follow the result in \cite{zhu2021one} to categorize all videos into 885 identities.

\textbf{Deepfake Detection (DFD).} DFD\cite{DFD} is a face swapping dataset released by Google, including 363 real videos of 28 actors and 3000 fake videos generated from them. The details of the face swapping methods involved in DFD are not available.

\textbf{CelebDF.} CelebDF\cite{li2020celeb} contains 590 real videos of 59 celebrities. And 5639 synthesized videos are generated by swapping faces between any two identities.

\textbf{DeeperForensics-1.0 (DFo).} DFo\cite{jiang2020deeperforensics} is a large-scale deepfake dataset featuring carefully captured real videos and a new face swapping method with fewer artifacts. 50000 real videos of 100 actors with specific expressions, lightning and different qualities are collected in this dataset. And 10000 fake videos are generated by swapping one of the 100 identities to one video in FF++.

Apart from the benchmark datasets above, we retrieve fake videos generated by more advanced face swapping methods, i.e., SimSwap\cite{chen2020simswap} and MegaFS\cite{zhu2021one}, to further evaluate the generalization to new methods. For MegaFS, we use their released fake images generated based on FF++. As for SimSwap, we generate fake videos with the released model by ourselves.

For each dataset, we train the face identification model with 10 real images per identity, which are randomly sampled from all the frames of the identity in the dataset. Then we sample 10,000 real and fake images to test our method and the frame-level AUC(\%) is used as the performance metric. It's worth noting that the real images in the training set are excluded from the test set.

\subsubsection{Implementation Details}
The face identification model is trained following the standard transfer learning pipeline. Specifically, we initialized the model with the widely used identity extractor ArcFace and added a fully connected layer to predict identities. In our baseline approach, the parameters of ArcFace are frozen. In the finetuning scheme against grey-box attack, the 4th stage and above are open in the IC\_SE50 ArcFace. Each model was trained for 30 epochs with a training batch size of 128. We use the Adam optimizer and the initial learning rate is set to 0.001, dropped by 0.1 after every 10 epochs.
As for the attention-based finetuning, the number of blocks in attention masks is set as 10, and the hyper-parameter $\alpha$ in label smoothing is set as 0.5. The model for generating gradient masks is updated every 5 epochs.

\subsection{Defense-unaware Forgery Detection}
In this section, we evaluate the face swapping detection performance of the proposed IdP\_FSD in the defense-unaware scenario. It means the fake images are generated without any knowledge about the detection method. We compare our approach with existing detection works on cross-dataset, cross-manipulation and cross-quality performance, and then analyze the influence of some factors inside our approach.

\subsubsection{Cross-dataset Performance}
\begin{table}[!t]
\renewcommand{\arraystretch}{1.3}
\caption{Cross-dataset Face Swapping Detection Result}
\label{table_cross_database}
\centering
    \begin{threeparttable}
    \begin{tabular}{m{2.6cm}<{\centering} m{0.9cm}<{\centering} m{0.8cm}<{\centering} m{1.3cm}<{\centering} m{0.8cm}<{\centering}}
        \toprule
        Method & FF++ & DFD & CelebDF & DFo\\
        \hline
        Xception\cite{rossler2019faceforensics++} & \textcolor{lightgray}{99.09} & 87.86 & 65.27 & -\\
        DSP-FWA\cite{Li_2019_CVPR_Workshops} & 74.82 & 80.59 & 64.87 & 45.46\\
        Face X-ray\cite{li2020face} & \textcolor{lightgray}{87.40} & 85.60 & 74.20 & -\\
        MultiAtt\cite{zhao2021multi} & \textcolor{lightgray}{99.29} & 79.01 & 67.44 & 89.58$^\ast$\\
        RFM\cite{wang2021representative} & \textcolor{lightgray}{98.79} & - & 65.63 & -\\
        LRL\cite{chen2021local} & \textcolor{lightgray}{\textbf{99.46}} & 89.24 & 78.26 & -\\
        RECCE\cite{cao2022end} & \textcolor{lightgray}{95.02} & 83.21 & 68.71 & 95.45$^\ast$\\
        DCL\cite{sun2022dual} & \textcolor{lightgray}{99.30} & 91.66 & 82.30 & -\\
        \hline
        ICT\cite{dong2022protecting} & 90.22 & 84.13 & 85.71 & 93.57\\
        ICT-Ref\cite{dong2022protecting} & 98.56 & \underline{93.17} & 94.43 & \underline{99.25}\\
        \hline
        IdP\_FSD & \underline{99.35} & \textbf{96.98}  & \textbf{99.33} & 99.21\\
        IdP\_FSD-ft & 98.69 & 92.69 & \underline{95.78} & \textbf{99.30}\\
        \bottomrule
    \end{tabular}
    \textbf{bold} and \underline{underline} denotes the best and second-best result of each dataset.
    \textcolor{lightgray}{Light-gray data} means intra-dataset evaluation for binary classification methods, i.e. training and testing on the same face forgery datasets.
    And $\ast$ represents the re-implemented result.
    \end{threeparttable}
\end{table}

Table \ref{table_cross_database} reports the face swapping detection performance across different datasets. We compare with two categories of state-of-the-art detection methods. The first category of methods models the forgery detection as a binary classification problem and is trained on real and fake faces. And the second category, i.e., ICT\cite{dong2022protecting}, is trained on real faces only, which is similar to our approach. ICT-Ref is the reference-assisted variant of ICT, whose reference set is built in the same way as our training set, i.e., randomly sampling 10 images for each identity in the dataset.

From the table, we can see that all the compared binary classification based methods suffer a performance drop on unseen datasets. On the other hand, our method achieves significant improvement and generalizes well on different datasets. Take CelebDF as an example, our method outperforms the listed binary classification based methods by over 17\% and also outperforms ICT-Ref with the similar setting by about 5\%. The improvement mainly benefits from exploiting the combination of identity features which applies to different face swapping forgeries. 

\subsubsection{Cross-manipulation Performance}
We further conduct a fine-grained experiment to demonstrate the generalization across different face swapping methods on FF++. Apart from the three face swapping methods provided by FF++, i.e., DF, FS and FSh, we include another two recent methods: SimSwap\cite{chen2020simswap} and MegaFS\cite{zhu2021one}, to evaluate our performance on new manipulations better. MegaFS-FF++ is made available by its original paper. As for SimSwap, we generate fake images with the same source and target real image pairs as other methods in FF++.

\begin{table}[!t]
\renewcommand{\arraystretch}{1.3}
\caption{Cross Manipulation Face Swapping Detection Result}
\label{table_cross_manipulation}
\centering
    \begin{tabular}{c c c c c c}
    \toprule
    Method & DF & FS & FSh & SimSwap & MegaFS\\
    \hline
    MultiAtt\cite{zhao2021multi} & \textcolor{lightgray}{99.51} & 67.33 & - & - & -\\
    RECCE\cite{cao2022end} & \textcolor{lightgray}{\underline{99.65}} & 74.29 & - & - & -\\
    DCL\cite{sun2022dual} & \textcolor{lightgray}{\textbf{99.98}} & 61.01 & - & - & -\\
    ICT\cite{dong2022protecting} & 94.20 & 90.87 & 79.70 & 78.31 & 95.20\\
    ICT-Ref\cite{dong2022protecting} & 99.10 & 98.76 & 97.45 & 95.83 & 99.85\\
    IdP\_FSD & 99.43 & \textbf{99.51} & \textbf{99.06} & \underline{98.37} & \underline{99.88}\\
    IdP\_FSD-ft & 99.12 & \underline{98.84} & \underline{98.66} & \textbf{98.50} & \textbf{99.93}\\
    \bottomrule
    \end{tabular}
\end{table}

As shown in Table \ref{table_cross_manipulation}, our approach outperforms the competitors on most manipulations. MultiAtt\cite{zhao2021multi}, RECCE\cite{cao2022end} and DCL\cite{sun2022dual} are among the state-of-the-art deepfake detection based on binary classifiers. When trained on Deepfakes and tested on FaceSwap, the AUC gaps are larger than 25\%. ICT\cite{dong2022protecting} is trained on real faces only and exploits identity inconsistency for face swapping detection, thus generalizes well on Deepfakes, FaceSwap and MegaFS. However, for new methods like FaceShifter and SimSwap which swap faces with less inconsistency, it also faces more than 10\% AUC drop. Moreover, when equipped with the same number of real faces of identities in FF++, our approach still outperforms ICT-Ref by about 2\% on SimSwap. Our approach avoids overfitting on specific forgery methods and makes use of a more common identity feature, thus improving generalization over different manipulations.

\subsubsection{Cross-quality Performance}
To evaluate the generalization ability to different image quality, we conduct experiments on the FF++ dataset with different compression levels. Image compression level is measured using JPEG quality factor (QF), where a smaller QF represents heavier compression. And we use the 
high-quality(c23) and low-quality(c40) FF++ to measure the performance on different video compression levels. 
The result AUC is reported in table \ref{table_compression}. MultiAtt\cite{zhao2021multi} and RECCE\cite{cao2022end} represent previous detection methods based on low-level features like textures. They are trained on FF++(c23) and evaluated on different compression levels. Both methods face a noticeable performances drop on QF=20 and the c40 compression level. Between them, MultiAtt shows better generalization because of augmentation.
On the contrary, our method, as well as ICT, makes use of high-level identity features, thus performing well even with heavy compression.

\begin{table*}[!t]
\renewcommand{\arraystretch}{1.3}
    \caption{Cross Quality Face Swapping Detection Result}
    \label{table_compression}
    \centering
    \begin{tabular}{m{3.5cm}<{\centering} m{1.3cm}<{\centering}m{1.3cm}<{\centering}m{1.3cm}<{\centering}m{1.3cm}<{\centering}m{1.3cm}<{\centering} | m{1.3cm}<{\centering}m{1.3cm}<{\centering}}
        \toprule
        \multirow{2}{*}{Method}  & \multicolumn{5}{c}{Image Compression (JPEG Quality Factor)} & \multicolumn{2}{c}{Video Compression} \\
        \cmidrule(lr){2-6} \cmidrule(lr){7-8}
        ~ & 90 & 70 & 50 & 30 & 20 & c23 & c40 \\
        \hline
        MultiAtt\cite{zhao2021multi} & 97.62 & 98.15 & 98.50 & 98.44 & 96.11 & 98.53 & 95.01 \\
        RECCE\cite{cao2022end} & 95.85 & 95.19 & 92.12 & 84.20 & 74.08 & 97.51 & 78.00 \\
        ICT-Ref\cite{dong2022protecting} & 98.13 & 98.15 & 98.10 & 98.01 & 97.90 & 98.16 & 98.19 \\
        IdP\_FSD & 99.35 & 99.35 & 99.34 & 99.33 & 99.22 & 99.35 & 98.04 \\
        IdP\_FSD-ft & 98.69 & 98.66 & 98.59 & 98.40 & 97.96 & 98.69 & 97.41 \\
        \bottomrule
    \end{tabular}
\end{table*}

\subsubsection{Training Set Sensitivity}
Since our approach requires real images of the concerned identities for training, it is necessary to analyze how sensitive the detection performance is to the training set selection. Facial variations such as pose, age and illumination are known to influence the performance of face related tasks\cite{tao2022frontal}. There is no doubt that increasing the number and variety of training real faces can boost the detection performance under challenging situations. However, it is common in the real world that only limited training images are available while the query images are various. Thus the low sensitivity to the difference between training and testing set is essential. In this section, we show using real images to train a face identification model is less sensitive than using them as a reference set.

\begin{table}[!t]
\renewcommand{\arraystretch}{1.3}
    \caption{Sensitivity to Pose Variance}
    \label{sensitivity_pose}
    \centering
    \begin{threeparttable}
        \begin{tabular}{c c c c c c }
            \toprule
             Test pose(real/fake) & F/P & F/F & P/P & P/F & Range $\downarrow$ \\
            \hline 
             ICT-Ref\cite{dong2022protecting} & 97.44 & 95.82 & 93.91 & 90.35 & 7.09 \\
             ArcFace-Ref & \textbf{98.85} & 97.43 & 97.81 & 93.73 & 5.12 \\
             IdP\_FSD & 98.55 & \textbf{97.99} & \textbf{98.46} & \textbf{97.32} & \textbf{1.23} \\
            \bottomrule
        \end{tabular}
        Here \textbf{F} indicates frontal faces and \textbf{P} indicates profile faces. The training set are all frontal faces.
    \end{threeparttable}
    
\end{table}

Take poses as an example, we design an experiment to study the sensitivity toward pose variance. For a face image, we estimate the angle of yaw and label the image whose absolute value of yaw is less than $10^{\circ}$ as frontal, and the image whose absolute value of yaw is larger than $25^{\circ}$ as profile.
The training set in our approach only contains 10 frontal images per identity while the testing set contains either frontal or profile images.
We take two methods that use real images as reference sets for comparison. 
As shown in table \ref{sensitivity_pose}, our approach has low sensitivity to pose variation. In particular, telling apart profile real faces and frontal fake faces is the most challenging setting. Under this setting, our approach outperforms the search-based competitors by 6.97\% and 3.59\% with the same requirement of reference images. It demonstrates that our approach is less sensitive.

\subsubsection{Identity Scalability}
\begin{figure}[!t]
\centering
\includegraphics[width=3.3in]{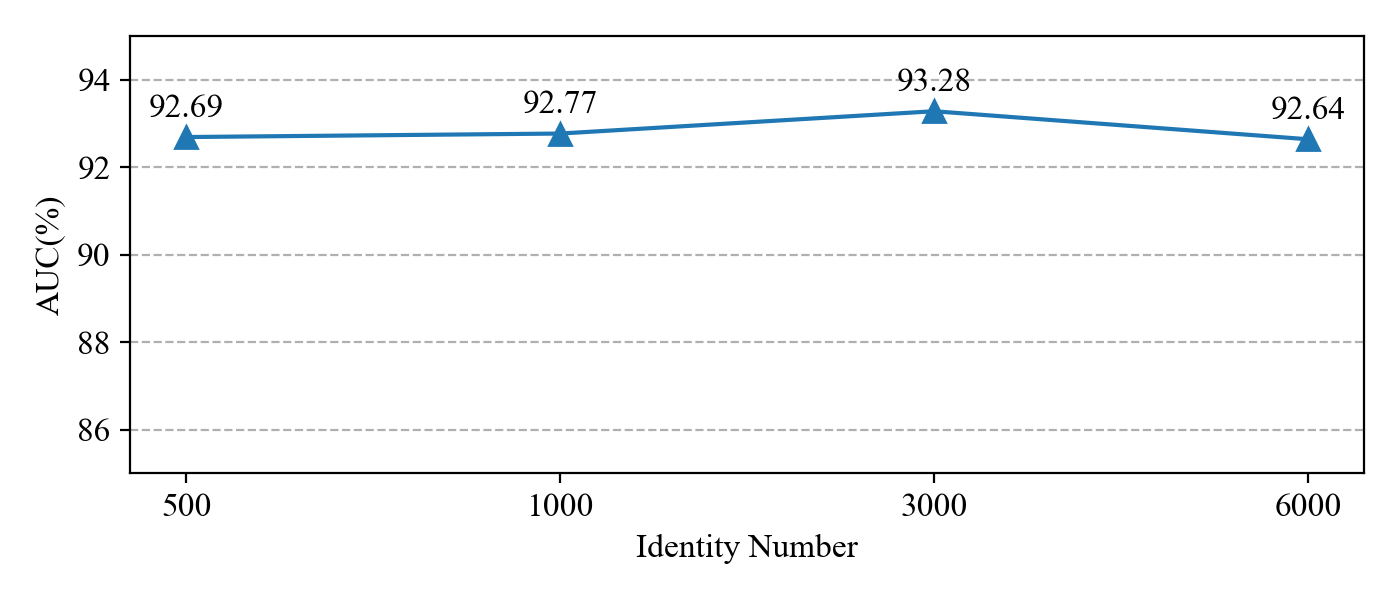}
\caption{The effect of the number of identities to be protected.}
\label{id_scalability}
\end{figure}

In this section, we evaluate how face swapping detection performance would change when the number of identities in the concerned set increases. The experiment is conducted on MegaFS\cite{zhu2021one} generated from CelebA-HQ, as there are about 6,000 real identities used to generate fake images, which is more than other datasets. 

The result is shown in figure \ref{id_scalability}. We train four face identification models with 500, 1,000, 3,000 and 6,000 identities respectively and report the AUC of using them in our detection method. For each model, the identities are randomly chosen from CelebA, as well as the real training images. And only the fake images whose source images belong to the chosen identities are included in the testing set. From the figure we can tell that the AUC hardly drops within the scale of 6,000 identities. Therefore, although our method can only detect face swapping images of a finite identity set, the identity set can scale to a large number.

In real-world applications, it is possible for the number of concerned identities to exceed thousands. In such situations, models like hierarchical classifier\cite{ma2018hierarchical} can be used to expand the scale of our approach. As a simulation, we divide the 6,000 identities into two groups with 3,000 identities each. Following our method, we train two face identification models. When detecting fake images, we send the test image to both models and get two maximum probability values. The larger one is used as the metric to detect swapped images. It turns out that using such two identification models achieves 92.73\% AUC, which is comparable to using a single model. In this way, our method can scale to an even larger identity set by dividing subsets.

\subsubsection{Analysis of Module Selection}

\begin{figure}[!t]
\centering
\includegraphics[width=3.0in]{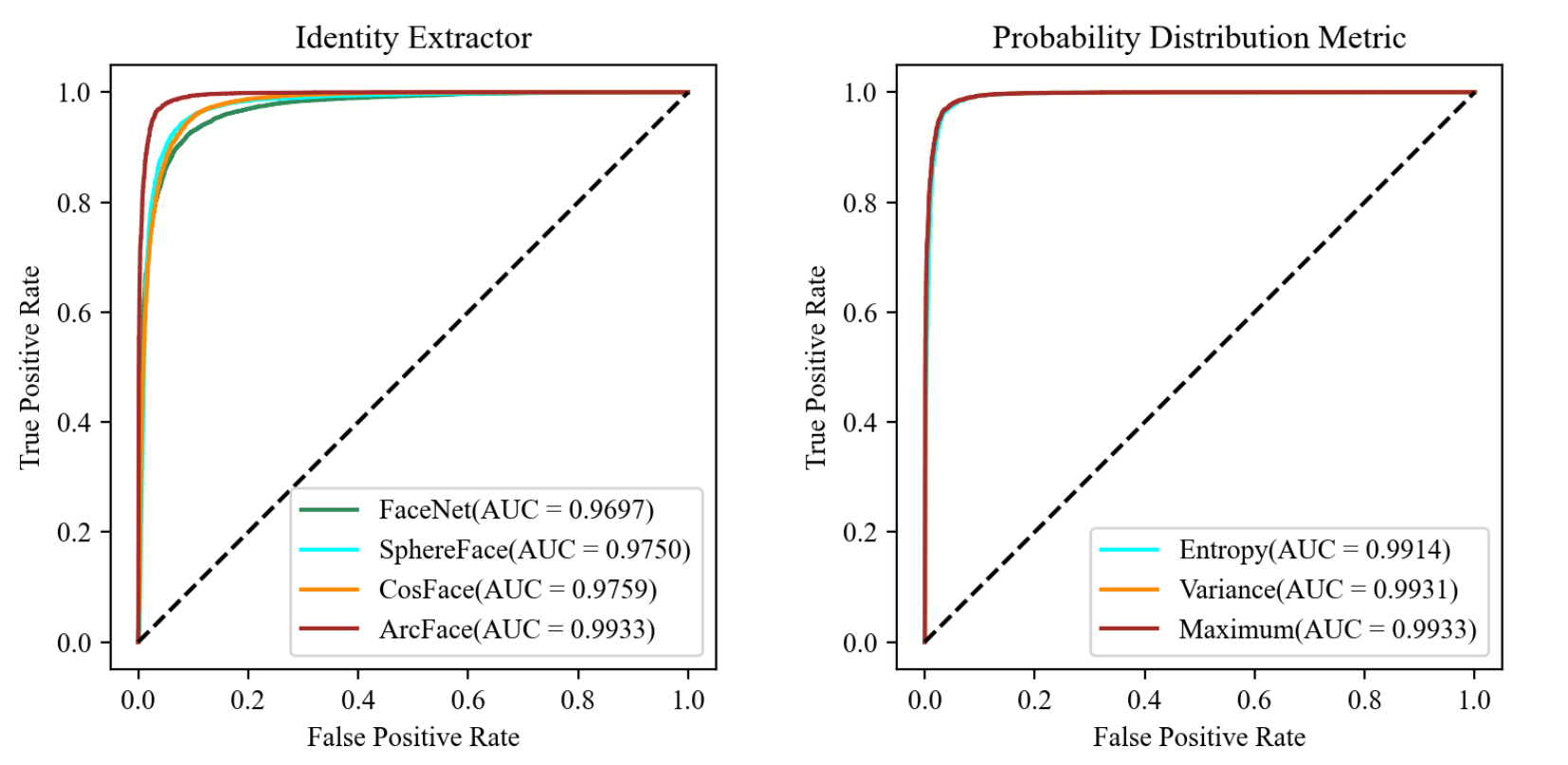}
\caption{Effect of different identity extractors (left figure) and metrics to measure the probability distribution (right figure) in terms of ROC curve and AUC at frame-level.}
\label{roc_curve_model_metric_selection}
\end{figure}

\textbf{The effect of identity extractor.} Here we study the effect of the identity extractor used in our approach. We evaluate the best publicly available face recognition models including FaceNet\cite{schroff2015facenet}, SphereFace\cite{liu2017sphereface}, CosFace\cite{wang2018cosface} and ArcFace\cite{deng2019arcface}. The comparison is conducted on CelebDF. As shown in the left figure of figure \ref{roc_curve_model_metric_selection}, ArcFace achieves the highest AUC. The reason might be Besides, AUC of the other three identity extractors all reach about 97\%, which means our face swapping detection pipeline applies to different identity extraction models.

\textbf{The effect of detection metric.} The maximum of the identification probability distribution is applied as our metric to detect face swapping images. Here we compare it with other metrics, i.e., the variance and entropy of the probability distribution. The experiment is carried out on CelebDF and the results are shown in the right part of figure \ref{roc_curve_model_metric_selection}. All three metrics perform well and the maximum metric is the best.

\subsection{Grey-box Evasion Attack Defending}

\begin{table*}[!t]
\renewcommand{\arraystretch}{1.3}
    \caption{Comparision of Grey-box Evasion Attack Defending Effectiveness}
    \label{defense_result_new}
    \centering
    \begin{tabular}{m{3cm}<{\centering} m{2cm}<{\centering} m{2.5cm}<{\centering} m{2.5cm}<{\centering} m{2.5cm}<{\centering}}
        \toprule
        \multirow{2}{*}{Method} & \multirow{2}{*}{AUC $\uparrow$} & \multicolumn{3}{c}{ASR $\downarrow$}\\
        \cmidrule{3-5}
        ~ & ~ & Single-ArcFace & Ensemble-3 & Ensemble-4 \\
        \hline
        IdP\_FSD & \textbf{99.33} & 99.46 & 46.56 & 98.60 \\
        Ensemble-4 min & 98.42 & 34.40 & 72.79 & 98.34 \\
        Ensemble-4 range & 97.11 & \textbf{21.72} & 59.84 & 96.72 \\
        Direct finetuning & 97.79 & 66.28 & 20.28 & 59.91 \\
        IdP\_FSD-ft & 95.78 & 44.75 & \textbf{8.93} & \textbf{37.64} \\        
        \bottomrule
    \end{tabular}
\end{table*}

\subsubsection{Effectiveness of Attention-based Finetuning Scheme}
In this section, we mainly study the effectiveness of the proposed attention-based finetuning framework in terms of defending the grey-box evasion attacks.
The threat model is defined in Section \ref{grey_box_threat_model}, i.e., the manipulator is aware of our detection mechanism, while the parameters are hidden from the manipulator. In such scenarios, the manipulator can launch evasion attacks by post-processing the swapped faces. And the goal of our finetuning scheme is to make it harder for a swapped face to evade the detection.

To simulate the evading process, we implement the attack described in Section \ref{grey_box_threat_model} using BIM attack\cite{kurakin2018adversarial} with an $l_\infty$ bound 4/255 and 20 iterations to find the $x_f^{adv}$ for each fake image $x_f$. Four publicly available identity extractors, i.e., FaceNet, CosFace, SphereFace and ArcFace, are used when minimizing the distance between the identities of the fake image and its source real image. Since our detector is implemented on ArcFace, we evaluate the following three model choices in the attack. The first is using a single identity extractor that is the same as the target detector, i.e., ArcFace. And the second is ensembling the other three extractors as in Eq. \ref{adv_eq}, to simulate the situation where the manipulator does not know the target architecture. Finally, we test ensembling all four extractors, which is practical since the manipulator may collect as many SOTA extractors as possible.

The attack success rate (ASR) is used to measure the effectiveness of defending schemes.
The ASR is calculated according to the EER threshold of 10,000 real and fake images without attack. Then the evasion attack is applied to the correctly detected fake images. The images whose maximum probability value becomes larger than the threshold are viewed as successfully evading examples. Except for ASR, keeping the defense-unaware detection performance is also important. Thus, we report the AUC without attack to measure the impact on normal performance. Table \ref{defense_result_new} shows the result of our method compared with several straightforward defense methods.

The baseline is to keep the identity extractor ArcFace frozen and train the classification layer only, shown in the first row. Despite the high detection AUC, this method is easily bypassed by evasion attacks. When the same identity extractor is exploited, almost all the fake samples can evade detection successfully. Even when the manipulator has no access to extractors with the same architecture, the ASR also reaches 46.56\%. That is to say, the attack noise that pulls a fake image and its real source close on some extractors can receive similar results on the unseen ones. Thus, the baseline method is vulnerable to evasion attacks and requires defense.

One straightforward defense is training multiple face identification models with different identity extractors and using the statics of their maximum probability values to detect fake images, denoted as 'Ensemble' in the table. In the experiment, we use the aforementioned four identity extractors and evaluate two statics which are minimum and range. From the table we can see that using multiple identity extractors can effectively defend the evasion attacks based on a single extractor. However, for the attacks that also use multiple extractors, the ensembling defense helps little in decreasing the ASR. Although including more different extractors may improve the defense, the same strategy works for the manipulator as well. 

Finally, we compare our attention-based finetuning scheme with direct finetuning. As the result shows, direct finetuning can defend part of the evasion attacks, since some attacks fail to transfer to the model with modified parameters. 
However, there are still about 60\% of attacks that successfully transfer and bypass the detection when the same architecture is included by the attacker. It is because without specification, identification models tend to focus on similar discriminative central areas. On the contrary, our finetuning scheme expands the attention area explicitly to further decrease the ASR to less than 40\%. 

\subsubsection{Justification of the Defense}
\textbf{Ablation Study.} This section verifies the effectiveness of the two parts in our attention-based finetuning scheme. The experiment is carried out on CelebDF. As shown in table \ref{defense_ablation}, both attention-based mask and label smoothing work in decreasing ASR and maintaining the  detection AUC. To be specific, only applying attention-based masks achieves the lowest ASR, while the AUC drops to about 90\%. The reason is that labeling the masked face with a one-hot label makes the identification model overconfident with parts of faces. As a result, some face-swapping images also get high probability maximums, making it more difficult to detect. With the help of smoothed labels, we keep the AUC at the SOTA level (see table \ref{table_cross_database} for comparison) and succeed in defending grey-box attacks at the same time.

\begin{table}[!t]
\renewcommand{\arraystretch}{1.3}
    \caption{Finetuning Scheme Ablation Study}
    \label{defense_ablation}
    \centering
    \begin{tabular}{cccc}
        \toprule
         Label Smoothing & Attention-based Mask & AUC $\uparrow$ & ASR $\downarrow$ \\
        \hline 
          & & 97.79 & 59.91\\
         \checkmark & & 97.25 & 52.29\\
          & \checkmark & 90.94 & 17.69\\
         \checkmark & \checkmark & 95.78 & 37.64\\
        \bottomrule
    \end{tabular}
\end{table}

\begin{figure}[!t]
\centering
\includegraphics[width=3.0in]{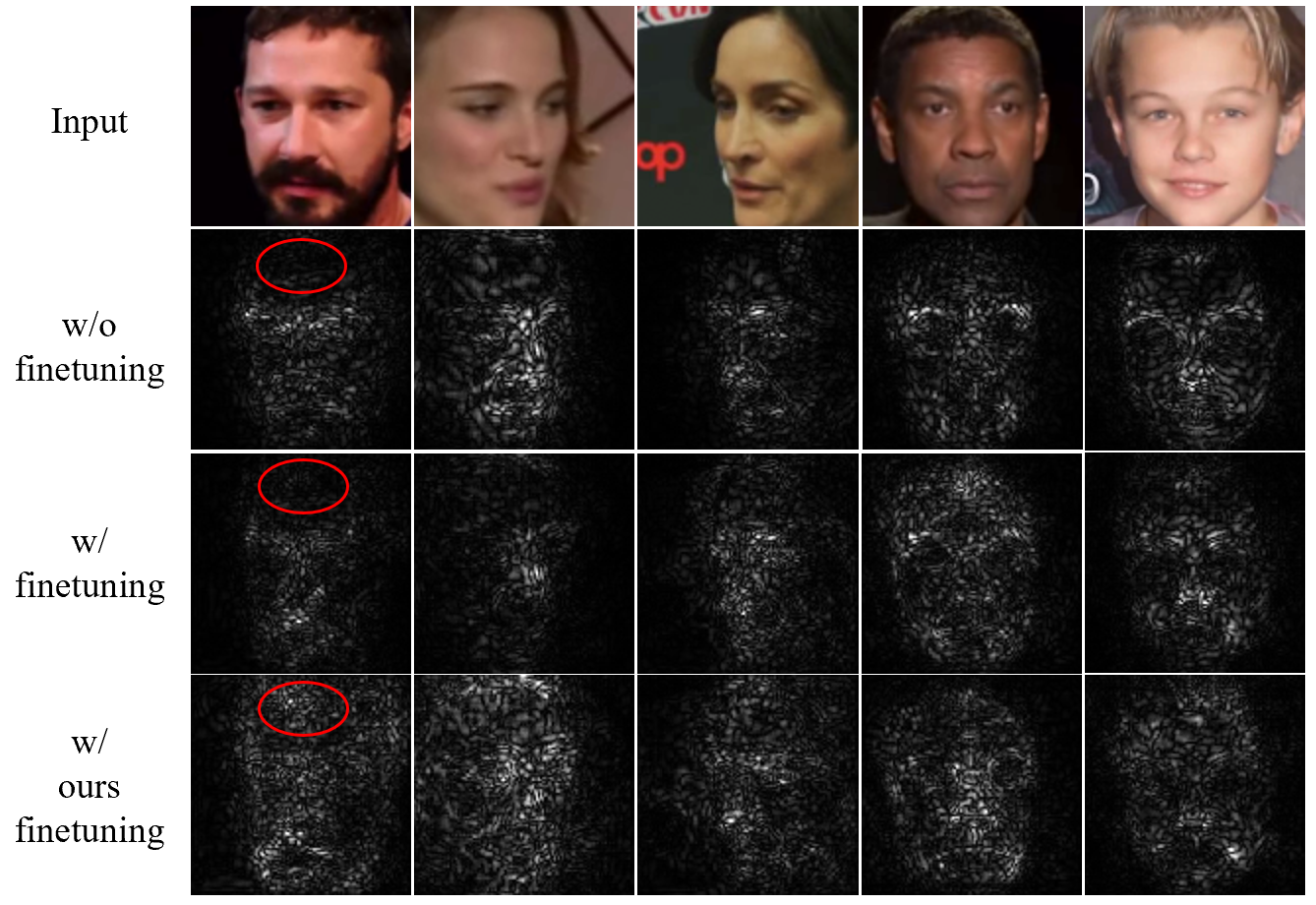}
\caption{Qualitative visualization of the attention map of the face identification model w/ and w/o the attention-based finetuning process. Lighter pixels indicate larger contributions to the predicted identity.}
\label{gradient_example}
\end{figure}
\textbf{Visualization.} 
To verify the effect of our attention-based finetuning scheme, we show the saliency maps of the face identification models with and without our finetuning in figure \ref{gradient_example}. We follow \cite{smilkov2017smoothgrad} to get the saliency map of the predicted identity based on gradients. The first row shows randomly selected real images sent to the face identification model. And the second row shows the corresponding saliency maps of the identification model trained on the frozen identity extractor, while the third row displays those of the model finetuned with our scheme on the partially open identity extractor. By comparing the saliency maps in two rows we can tell that the attention area is expanded by our finetuning scheme. The frozen identity extractors mainly pay attention to the central face area. On the contrary, for the finetuned identification model, the whole face contributes to the identity classification. For example, the red circle in the first column highlights the attention change in the forehead area. As a result, the evasion attacks conducted on the frozen identity extractors are more difficult to transfer to our finetuned identification model.

\begin{figure}
    \centering
    \includegraphics[width=3.0in]{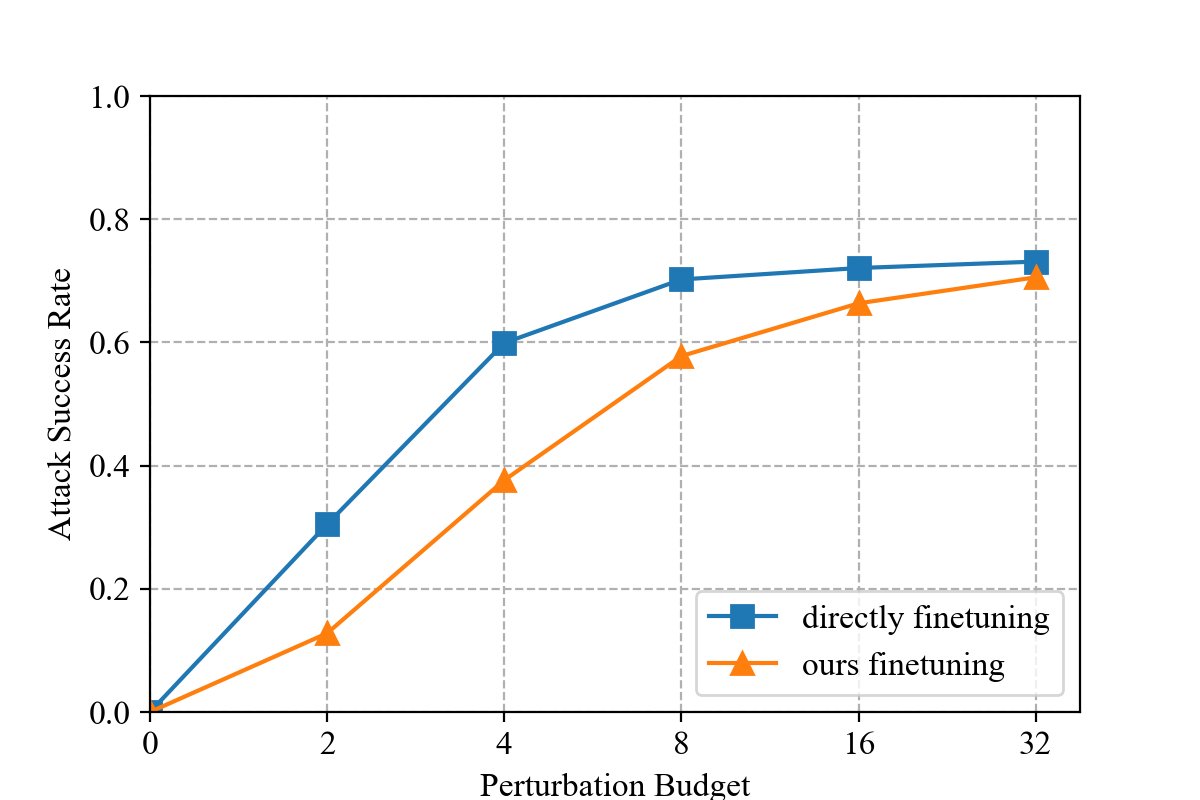}
    \caption{Effects of perturbation budgets under the $l_\infty$ norm on ASR.}
    \label{attack_epsilons}
\end{figure}
\begin{figure}
    \centering
    \includegraphics[width=3.0in]{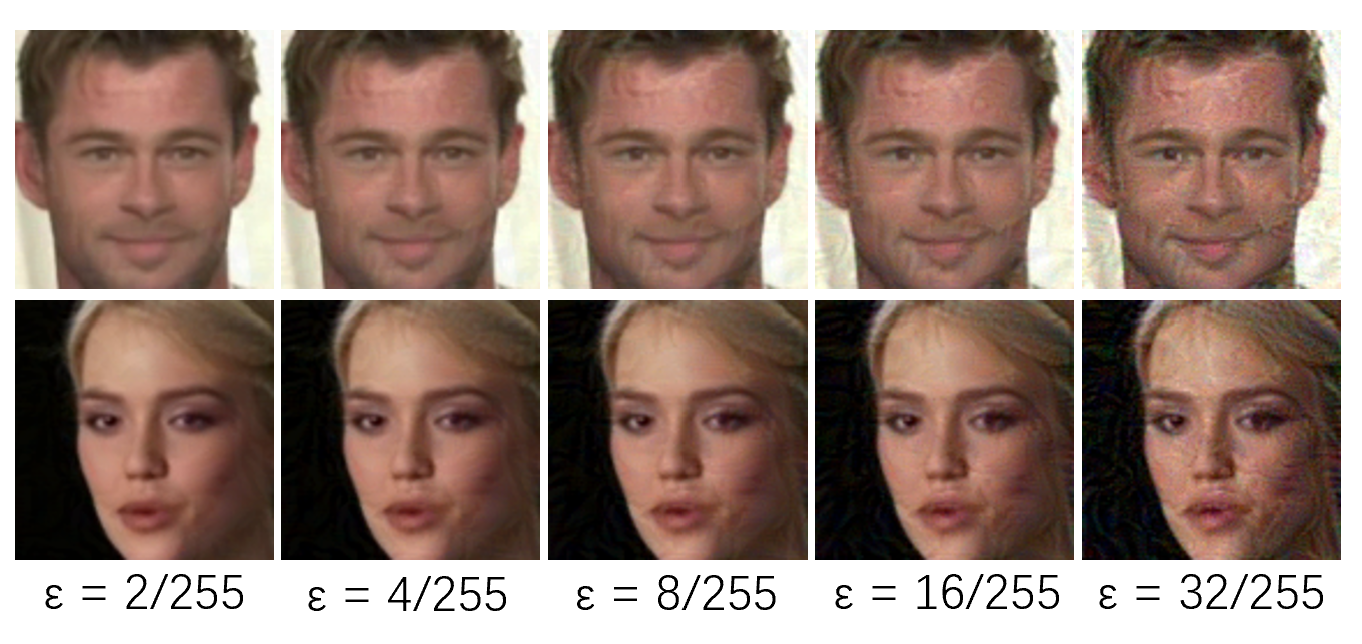}
    \caption{Effects of perturbation budgets under the $l_\infty$ norm on visual quality.}
    \label{attack_visual_quality}
\end{figure}
\textbf{Effects of Perturbation Budgets.}
Increasing the perturbation budget can boost the ASR while lowering the visual effects of the swapped faces. We report such effects in Fig \ref{attack_epsilons} and Fig \ref{attack_visual_quality}. As shown in the Fig \ref{attack_epsilons}, the attack success rate goes up to about 70\% when the $l_\infty$ norm bound is set to 32/255. However, the adversarial noise becomes obvious to human eyes under this perturbation bound, as shown in the rightmost column in Fig \ref{attack_visual_quality}. Meanwhile, visual quality is especially important for face swapping images which aim at deceiving human eyes.
Therefore, it is reasonable to use a smaller budget 4/255 used in our experiment to evaluate the defense performance, considering the invisibility.

\textbf{Adaptive Attacks.} This section explores the finetuning defense's robustness when the manipulator also knows the existence of our finetuning scheme. To attack IdP\_FSD-ft in a straightforward way, the manipulator might train a surrogate face identification model with the same pipeline for transfer attack.
We assume the manipulator cannot access our training set, including the images as well as the identity set. What the manipulators can have are images with the identity they want to swap. That's to say, the training set of the manipulator has overlap with that of the detector but is not exactly the same.

In particular, we randomly select 10 identities from CelebDF as the source identities of the manipulator. The surrogate model is trained on these 10 identities with the same attention-based finetuning approach, while the detector to be attacked is trained on all 59 identities in CelebDF. Then with the trained surrogate, the manipulator conducts a BIM attack to increase the swapped image's maximum probability on the surrogate.

\begin{table}[!t]
\renewcommand{\arraystretch}{1.3}
    \caption{Robustness of IdP\_FSD-ft against Adaptive Attack}
    \label{defense_threat_model2}
    \centering
    \begin{tabular}{m{3.8cm}<{\centering} m{1.6cm}<{\centering} m{1.6cm}<{\centering}}
        \toprule
         Method & AUC $\uparrow$ & ASR $\downarrow$ \\
        \hline 
         Direct finetuning & 97.79 & 62.55\\
         IdP\_FSD-ft & 95.78 & 58.39\\
        \bottomrule
    \end{tabular}
\end{table}

We report the attack result in table \ref{defense_threat_model2} and compare it with the direct finetuning method in the same setting. Under the challenging adaptive attack situation, our method still defends about 40\% of the attacks. Moreover, the attention-based finetuning scheme outperforms direct finetuning.
The result shows that our defense raises the bar of evasion attacks. However, the relatively high ASR also tells that tackling adaptive attacks is still an open problem and can be improved in future work.

\section{Discussion and Limitation}
\textbf{Other Face Forgery Types.} Our detection method is specially designed for face swapping, one of the various face forgery types. We make use of the nature of face swapping, i.e., the swapped face inevitably combines the identities of the two faces involved in the swapping. Thus, for those forgery types without identity swap like face reenactment, our method cannot work. Considering the variety of face forgeries, it is difficult to detect them with a single method\cite{dong2022protecting}. In addition, since face swapping is rather popular and has developed lots of different methods, we think it is meaningful to study detectors that generalize well among face swapping methods. 

\textbf{Image and Video Level Detection.} We focus on face swapping image detection and the comparison is also done with the SOTA image-level detection methods. To detect fake videos, we can aggregate the predictions of all or some frames, as in \cite{haliassos2021lips}. On the other hand, there are a number of detectors specially designed for videos that perform well, utilizing inter-frame consistency\cite{gu2022delving}, temporal correlation\cite{hu2022finfer} and so on. One related work is ID-Reveal\cite{cozzolino2021id}, which learns identity features from face landmark sequences and compares a probe video with reference real videos. ID-Reveal mainly targets face reenactment video detection, while our IdP\_FSD focuses on face swapping images. Thus, we regard our image-level detection methods and the video-level ones like ID-Reveal as complements, as both forgery images and videos can spread and cause malicious effects.

\section{Conclusion}
This paper presents a generalizable and robust face swapping detection approach by utilizing the identity combining nature of face swapping. The approach trains a face identification model on real images and detects swapped faces according to how confused the identification model is. We measure the confusion with the maximum of the probability distribution. To defend against grey-box attacks, an attention-based finetuning scheme is proposed to expand the attention area of the identification model. Our approach is designed for detecting fake images for a specific set of identities, which is common in real-world applications. 

We evaluate our detection method on several benchmark datasets. In the defense-unaware scenario, extensive experiments show our method achieves high detection performance on different face swapping methods. Compared with existing detection works, we get better generalization cross manipulation and image quality. In the grey-box scenario where the manipulators post-process the fake images to bypass our detector, the proposed finetuning scheme decreases the evasion rate. 

\ifCLASSOPTIONcaptionsoff
  \newpage
\fi



\bibliographystyle{IEEEtran}
\bibliography{reference.bib}
\end{document}